\RequirePackage{fix-cm}
\documentclass{svjour3}                     % onecolumn (standard format)
\smartqed  % flush right qed marks, e.g. at end of proof
\usepackage{graphicx}
\usepackage{mathptmx}      % use Times fonts if available on your TeX system
\usepackage{algorithm}
\usepackage{algorithmicx}
\usepackage[noend]{algpseudocode}
\usepackage{amssymb}
\usepackage{enumitem}  % roman sub items
\usepackage{graphicx}
\usepackage[utf8]{inputenc}
\usepackage{mathtools}
\usepackage[numbers]{natbib} % fix bib style (natbib) for bib tex use
\usepackage{siunitx}
\usepackage{subfigure}
\usepackage{tabularx}
\usepackage{textcomp}
\usepackage{tikz}
\usetikzlibrary{decorations.pathreplacing}
\usepackage{url}
\usepackage{xcolor}
\newcommand{\SEmath}{SE(3)}
\newcommand{\SE}{$\SEmath$}
\newcommand{\SEdefinitionmath}{\SEmath=\mathbb{R}^3\times SO(3)}
\newcommand{\SEdefinition}{$\SEdefinitionmath$}

\newcommand{\MyAlgs}{$\kappa$-RRT-Connect}
\newcommand{\MyAlgA}{$\kappa$-SB-RRT-Connect}
\newcommand{\MyAlgB}{$\kappa$-B-RRT-Connect}
\newcommand{\MyAlgsUni}{$\kappa$-RRT-Connect}

\newcommand{\AlgSBRRTfull}{Spline-Based-RRT}
\newcommand{\AlgBRRTfull}{Bevel-Tip-RRT}

\definecolor{veinColor}{RGB}{0,0,255}
\definecolor{arteryColor}{RGB}{255,0,0}
\definecolor{cochleaColor}{RGB}{0,255,0}
\definecolor{semiColor}{RGB}{255,170,0}
\definecolor{ossicleColor}{RGB}{170,0,255}
\definecolor{chordaColor}{RGB}{85,255,255}
\definecolor{facialColor}{RGB}{255,255,127}
\definecolor{IACColor}{RGB}{255,0,255}
\definecolor{EACColor}{RGB}{170,170,127}

\newcommand{\ColorBox}[1][red]{
	\textcolor{#1}{\rule[0.1ex]{4pt}{1.1ex}}
}

\newcommand{\introLegend}
{
	\ColorBox[veinColor] jugular vein 
	\ColorBox[arteryColor] carotid artery 
	\ColorBox[facialColor] facial nerve
	\ColorBox[chordaColor] chorda tympani
	\ColorBox[EACColor] external auditory canal
	\ColorBox[IACColor] internal auditory canal
	\ColorBox[cochleaColor] cochlea
	\ColorBox[semiColor] semicircular canals
	\ColorBox[ossicleColor] ossicles
}
\journalname{IJCARS}
\begin{document}

\title{Planning Nonlinear Access Paths for Temporal Bone Surgery%\thanks{Grants or other notes
}


\author{Johannes Fauser         \and
		Georgios Sakas \and
		Anirban Mukhopadhyay
}


\institute{Johannes Fauser, Georgios Sakas, Anirban Mukhopadhyay
%	Georgios Sakas \\
%	Anirban Mukhopadhyay
	\at
	Department of Computer Science, Technische Universität Darmstadt, Germany\\
%	Technische Universität Darmstadt\\
%	Germany\\
%    Tel.: +49 6151 155-427\\
    \email{johannes.fauser@gris.tu-darmstadt.de}           %  \\
%            \emph{Present address:} of F. Author  %  if needed
 %          S. Author \at
  %            second address
}

\date{Received: date / Accepted: date}

\maketitle

\begin{abstract}
\textbf{\newline \\* Purpose:}
Interventions at the otobasis operate in the narrow region of the temporal bone where several highly sensitive organs define obstacles with minimal clearance for surgical instruments.  Nonlinear trajectories for potential minimally-invasive interventions can provide larger distances to risk structures and optimized orientations of surgical instruments, thus improving clinical outcomes when compared to existing linear approaches. 
In this paper, we present fast and accurate planning methods for such nonlinear access paths. \smallskip
\textbf{\\* Methods:} 
We define a specific motion planning problem in \SEdefinition~with notable constraints in computation time and goal pose that reflect the requirements of temporal bone surgery. 
We then present \MyAlgs: two suitable motion planners based on bidirectional Rapidly-exploring Random Trees (RRT) to solve this problem efficiently.\smallskip
\textbf{\\* Results:} 
The benefits of \MyAlgs~are demonstrated on real CT data of patients. Their general performance is shown on a large set of realistic synthetic anatomies. We also show that these new algorithms outperform state of the art methods based on circular arcs or Bézier-Splines when applied to this specific problem.\smallskip
\textbf{\\* Conclusion:} 
With this work we demonstrate that pre- and intra-operative planning of nonlinear access paths is possible for minimally-invasive surgeries at the otobasis.

%\keywords{First keyword \and Second keyword \and More \and at least for, up to six}

%% 200-300 words !!!!
\end{abstract}

\section{Introduction}
\label{sec:intro}
\begin{figure}[h]
	\centering
	\flushleft\introLegend
	\resizebox{\linewidth}{!}{%
		\includegraphics[height=1cm]{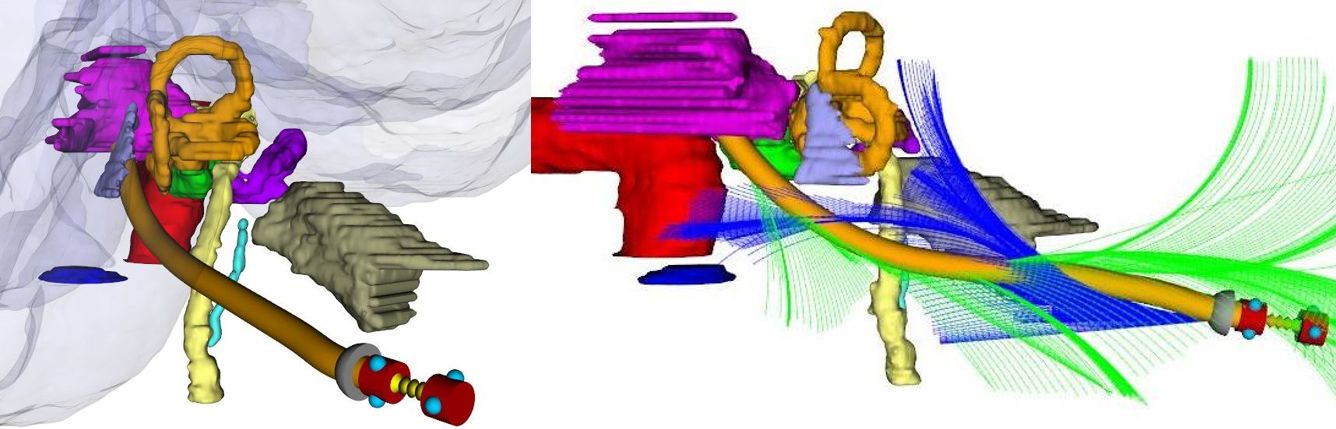}
	}
	\caption{
		A nonlinear access path (orange) has to be drilled through the temporal bone (empty space) by a surgical robot (colored) to reach the target of an intervention. 
		Various organs (different colors) form obstacles that our planning algorithms have to avoid (e.g. blue and green search graph of a RRT-Connect).
	}
	\label{fig:overall}
\end{figure}

In the last decades more and more minimally invasive procedures are introduced in the clinical work place \cite{beasley2012:review}. 
At the otobasis, the focus of research has been the drilling of either a single \cite{Gerber2014:PlanningTool,noble2010:linearDrilling} or multiple \cite{stenin2014:multiportLinear} linear access paths through the temporal bone to the cochlea and initial reports on clinical studies have been presented \cite{labadie2014:firstReport,caversaccio2017:firstReport}. 
In such interventions several obstacles or risk structures, e.g. the facial nerve and its small branch (Fig. \ref{fig:overall}, yellow objects), severely limit the space that is available for drilling. 

Unlike these linear approaches, nonlinear drilling provides several potential advantages: larger distances to risk structures, correcting misalignments while drilling, and optimization of orientation at the goal point (e.g. for the insertion of the electrode during a cochlear implantation or for instrument alignment). 
Yet, nonlinear planning at the otobasis is difficult to deploy due to the limited space and time constraints on intra-interventional planning. 
To the best of our knowledge, such an approach has never been investigated. 
In this paper, we consider the use of a curvature constrained drilling unit and propose two new RRT-Connect \cite{lavalle2006:book} algorithms to quickly (re-)compute feasible access paths for said robot: 
once at the beginning of the intervention and regularly in between if navigation errors occure (Fig. \ref{fig:overall}).

In recent years, significant progress has been made in the development of continuum robots and instruments for minimally invasive medical applications \cite{Burger2015:Survey}. 
Many of these can be categorized as "curvature constrained objects". 
These include, for example, steerable needles \cite{duindam2008:needles,cowan2011:needleSteering} for interventions in soft tissue \cite{patil2014:rapidReplanning,engh2006:deepBrain} or flexible endoscopes \cite{fichera2017:ThroughTheTube}. 
If such underactuated systems are used, where instrument steering is limited to certain directions, nonholonomic motion planning based on the Rapidly-Exploring Random Tree \cite{lavalle2006:book,Alterowitz2008:book} or even optimal randomized motion planners \cite{karaman2012:holonomicAlgorithms,gammell2014:BIT} are used to plan feasible trajectories around obstacles for the underlying instrument.

In medical applications the main focus mainly lies on steerable needles where planning is done with variants of the RRT. These methods consider, for example, special distance functions \cite{patil2014:rapidReplanning} or the reachable set of the nearest states \cite{shkolnik2009:reachabilityRRT}. Other methods speed up the convergence via potential fields \cite{yang2016:GART} or utilize heavy parallelization \cite{liu2016:fractalTree}.

The planning of trajectories for (unmanned) aerial vehicles such as drones or missiles also requires curvature constrained motion planning. Here, the development of an analytical solution of the 3D Dubins Problem \cite{hota2010:Optimal3D} lead to an RRT*-solver \cite{pharpatara2017:trajectoryRRTstar} in the case that start and goal regions are sufficiently far away. Moreover, Yang et al. \cite{yang2014:SBRRT,yang2014:optimalBezier} presented an RRT* with Bézier splines as local planning technique. 

However, curvature constrained motion planning for temporal bone surgery requires fast and precise algorithms with start and goal regions in \SE~within a small and dense environment. Although we were able to show the general feasibility of such trajectories within the otobasis \cite{fauser2016:softwareTool}, a reliable method does not yet exist. 

The main contribution in this paper is then twofold: 
First of all, we present two RRT-Connect algorithms which achieve fast path planning for nonlinear temporal bone surgery. 
%Secondly, we present a novel evaluation of such planners in synthetic anatomies based on statistical shape models from real CT patient data.
Secondly we address a novel evaluation strategy in case of limited annotated data sets: 
The robustness of such planners is shown on synthetic anatomies based on statistical shape models from real CT patient data.

\section{Objective}
\label{sec:objective}
Minimally-invasive procedures require a planning step that computes feasible trajectories while respecting potential constraints such as clearance to organs or instrument mobility. After the computation of a set of solutions these are then optimized according to a cost function or other optimization strategies \cite{stenin2014:multiportLinear,hamze2017:evolutionaryDeepBrain}.

In motion planning relevant parameters are usually expressed in a specific Problem Formulation \cite{lavalle2006:book}.
In this section we describe the details for temporal bone surgery and how they are incorporated in the following Formulation:
\newline
\\*\noindent\textbf{Problem Formulation For Temporal Bone Surgery}
\begin{enumerate}
	\label{form:probDef}
	\item Let $O\subset\SEdefinitionmath$ be the obstacle region, defined by the location of several risk structures $\{\mathcal{R}\}_i\subset\mathbb{R}^3, 0\leq i\leq N$. 
	I.e. $O := \{q=(x,h)\in\SEmath \vert \exists i, 0\leq i\leq N : x\in\mathcal{R}_i\}$. 
	Let $C_{free} = \left\{q\in\SEmath\vert q\notin O \right\}$ be the free space of the configuration space.
	\item Let $C_I\subset C_{free}$ be the initial region.
	\item Let ${M_G\subset C_{free}}$ be a set of states. The goal region $C_G$ is then defined as
	\begin{multline*}
	C_G \equiv C_G(\epsilon_G, \phi_G) = \{ q(x,h)\in \SEmath~|\notag \\ \Vert x-y\Vert_{\mathbb{R}^3}<\epsilon_G, \rho(h,g)<\phi_G, \text{for a}~\hat{q}(y,g)\in M_G\},
	\end{multline*}
	where $\rho$ is defined as in Equation \ref{eq:norm} and
	\begin{enumerate}[label=(\roman*)]
		\item ${\epsilon_G \in \mathbb{R}^+}$ is the maximally allowed Euclidean distance and
		\item $\phi_G\in [0, \pi]$ is the maximally allowed angular difference at a specific goal state.
	\end{enumerate}
	\item Let ${d_{max}\in\mathbb{R}^{0+}}$ be the safety distance to risk structures. Let $r_d\in\mathbb{R}^+$ be the radius and ${\kappa_{max}\in\mathbb{R}^+}$ the maximum curvature constraint of the drilling robot.
	\item Let $T_{max}\in\mathbb{R}^+$ be the maximum time constraint available for planning.
	\item Task: Find a path $\gamma(t) : [0,1] \rightarrow \SEmath$ satisfying 
	\begin{enumerate}[label=(\roman*)]
		\item $\gamma(0)\in C_I$
		\item $\gamma(1)\in C_G$
		\item $\forall t\in(0,1): \left\Vert\gamma''(t)\right\Vert < \kappa_{max}$
		\item $\forall t\in[0,1], o\in O: \left\Vert\gamma(t)-o\right\Vert_{\mathbb{R}^3} > r_d + d_{max}$
	\end{enumerate}
	or report that no path could be found in the available time $T_{max}$.
\end{enumerate}

\textbf{Item 1} of this Problem Formulation introduces obstacles in $\mathbb{R}^3$ (e.g. the facial nerve) that have to be circumnavigated, as well as the free space, which defines potential positions the drilling unit can occupy.
\textbf{Item 2} corresponds to potential positions at the skull's surface that serve as entry points of instruments, whereas \textbf{Item 3} defines a spherical volume in $\mathbb{R}^3$ as the intervention's target together with a threshold $\phi_G$ that limits the potential orientation within this target volume. 
The orientation between two configurations is compared in the quaternion metric (see e.g. \cite{lavalle2006:book})
\begin{align}
\label{eq:norm}
\begin{split} % to get single ref number for whole equation
\rho(h_1, h_2) &= \text{min}\left\{\rho_S(h_1,h_2), \rho_s(h_1,-h_2)\right\}\\
\rho_s(h_1,h_2) &= cos^{-1}(a_1a_2+b_1b_2+c_1c_2+d_1d_2).
\end{split}
\end{align}

\begin{figure}[t]
	\centering
	\resizebox{\linewidth}{!}{%
		\includegraphics[height=1cm]{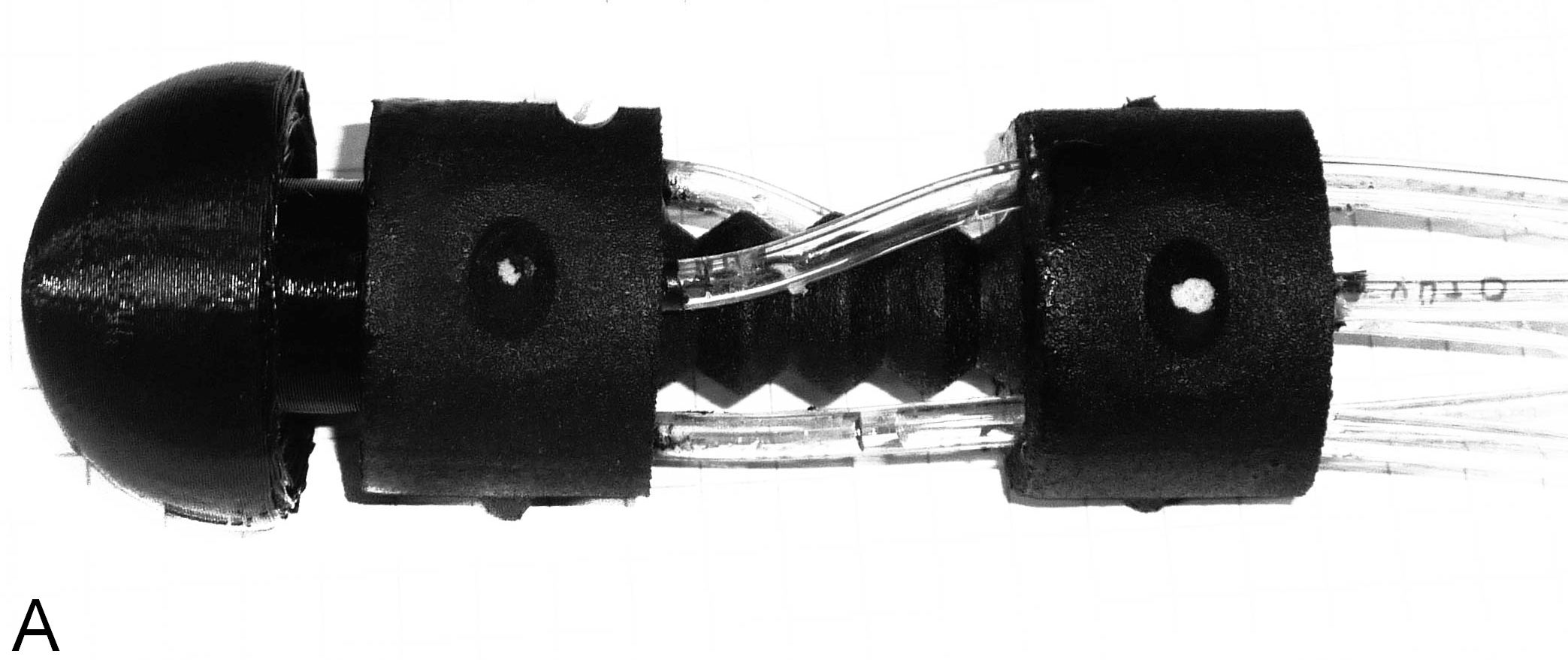}
		\includegraphics[height=1cm]{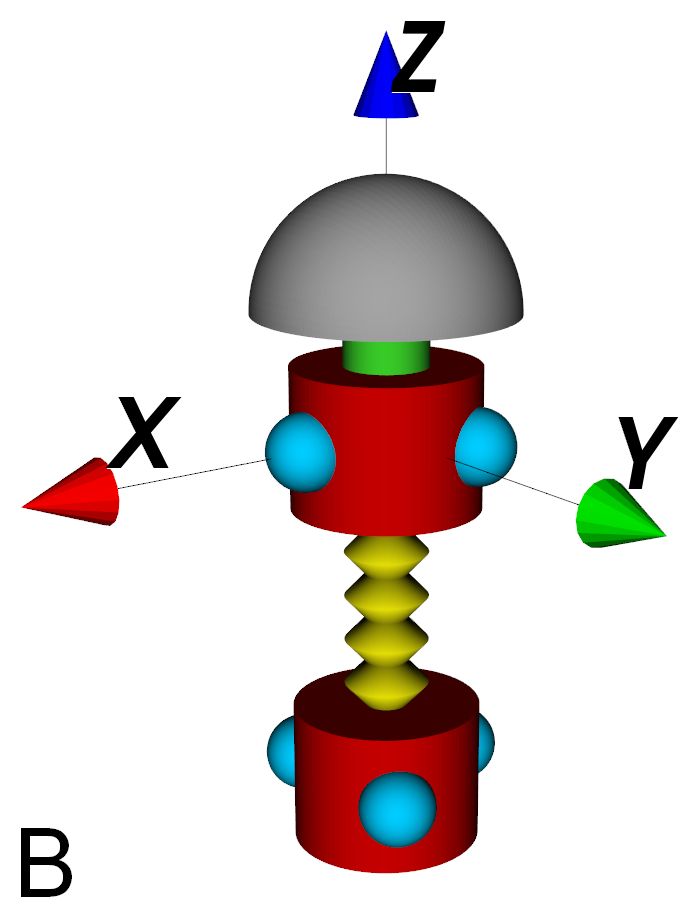}
	}
	\caption{\textbf{A:} Prototype of the hydraulically driven robot with a drill bit at the front. 
		\textbf{B:} Model of the robot based on geometric primitives and its local coordinate frame.}
	\label{fig:robotCloseUp}
\end{figure}
In this paper, we consider the use of a prototype robot (Fig. \ref{fig:robotCloseUp}) currently under development for the creation of nonlinear access paths. 
If we consider the z-Axis in the local coordinate frame of the model as the robot's \textit{line of view}, we want to match its pose with the ones at start and goal states: 
Initial states should be close to the skull's surface normal in order to minimize deviation from the desired trajectory due to forces applied during drilling.
For a cochlear implantation, for example, goal states at the round window would represent the optimal insertion angle. Here, work has been done to limit the deviation from the optimum to less than $\ang{5}$ \cite{torres2017:improvementCochleaAngle}. 

The robot's limitations are then included via \textbf{Item 4}: 
The radius of the drill bit and an additional safety distance to account for navigation errors or heat generation are combined to a distance constraint. 
Additionally, the maximum turning angle of the prototype results in a curvature constraint. 
\textbf{Item 5:} Potential misalignments during navigation require an intra-interventional replanning step to either provide a new corrected trajectory or stop the drilling. 
Therefore, an algorithm has to be fast enough to provide a smooth intervention. 
\textbf{Item 6:} A motion planning algorithm for this procedure will then try to find a feasible path in the available time which would result in a trajectory connecting both a start and a goal state (i, ii), observing a maximally allowed curvature (iii) and last, a necessary distance to risk structures (iv).

Note 1: This formulation remains valid in the case of replanning where the initial region $C_I$ of Item 2 will be set to the current pose of the robot.

Note 2: This formulation extends the problem of trajectory planning in soft tissue for bevel-tip needles, where alignment of instruments \cite{schulman2014:convexOptimization} and regular fast replanning \cite{patil2014:rapidReplanning} is needed, by introducing constraints on both start and goal orientations. We expect our planners to be useful for this kind of application as well.

\section{Methods}
\label{sec:methods}
The main difficulty of this problem is the fast and precise matching of the goal's pose. 
An intuitive way to address this problem is to use an RRT-Connect algorithm \cite{kuffner2000:RRTconnect}. 
This method, unlike basic RRTs, grows search trees from both the goal and the initial region in an attempt to connect these two. 
With this strategy more possible connections are available than just those between search tree and goal regions. Thus, successfully finding an access path is more likely. 
The general RRT-Connect can be described as follows (Algorithm \ref{algo:k-RRT-Connect}):

\newcommand{\TreeI}{\mathcal{T}_I}
\newcommand{\TreeG}{\mathcal{T}_G}
Two trees $\TreeI, \TreeG$ are initialized with states of the initial and the goal region, respectively. 
Both trees are iteratively extended until either the maximally allowed time $T_{max}$ is reached or the graphs are connected successfully. 
In each iteration the two search trees take turn in the following procedure: 
A random state is drawn from the free space $C_{free}$. 
Then, the nearest neighbors to the current tree are computed according to a previously defined distance function. 
For each of these configurations the local steering function computes an expansion towards the random state. 
If no collision with obstacles occurs along this path the state is added to the tree. 
Last, the algorithm tries to connect both trees according to the state space's constraints (in our case the path needs to be two times continuously differentiable). 
If both trees are connected within the given time threshold $T_{max}$ the resulting path is returned. 
Otherwise, failure is reported.
\newcommand{\Tree} {\mathcal{T}}
\begin{algorithm}[h]
	\caption{\MyAlgsUni}
	\label{algo:k-RRT-Connect}
	\begin{algorithmic}[1]
		\State $\TreeI \gets \text{initial\_states}()$
		\State $\TreeG \gets \text{goal\_states}()$
		\While{ time\_spend$() < T_{max}$ and $not\_connected(\TreeI, \TreeG)$}
			\State $q_{rand} \gets$ sample\_state($C_{free}$)
			\State $\Tree \gets$ alternate($\TreeI, \TreeG$)
			\State $\{q\}_k \gets$ k\_nearest\_neighbors$(\Tree, q_{rand})$
			\ForAll{$q_{near}$ in $\{q\}_k$}
				\State $ q_{next} \gets \text{steer}(q_{near}, q_{rand}, \Delta t)$
				\If {collision\_free$(q_{near}, q_{next})$}
					\State extend\_tree$(\Tree, q_{near}, q_{next})$
				\EndIf
			\EndFor
			\State \textbf{end for}
			\State attempt\_connection$(\TreeI, \TreeG)$
		\EndWhile
		\State \textbf{end while}
		%		\EndProcedure
	\end{algorithmic}
\end{algorithm}

In the following, we shortly recall two local steering methods, one based on circular arcs of varying curvature and one based on Bézier-Splines. 
We then present two individual solutions that extend these planners to RRT-Connect versions.

\noindent\textbf{\AlgBRRTfull~\& \MyAlgB:} 
We use the local steering function developed for Bevel-Tip needles presented in \cite{patil2014:rapidReplanning} to create access paths of variable curvature. 
This method extends the search tree along circular arcs of variable radii. 
The RRT-Connect version uses Dubins Paths in 3D to connect the search trees as, unlike circular arcs, this is a technique to connect states in \SE.

\noindent\textbf{\AlgSBRRTfull~\& \MyAlgA:} 
The second RRT utilizes cubic Bézier-Splines to interpolate in \SE, resulting in an approximation of states in the search tree and a two times continuously-differentiable trajectory \cite{yang2014:SBRRT}. 
Here, the local steering method can be used naturally to attempt a connection. 

\noindent The individual steps in Algorithm \ref{algo:k-RRT-Connect} (lines 4,6,8,12) %for the \MyAlgs~ 
are then as follows:

\noindent\textit{sample\_state}: 
Sampling in \SE would require solving a two point boundary value problem, i.e. matching both location and orientation at the random state. 
This is not possible with either steering function. 
Instead, a state is merely sampled in $\mathbb{R}^3$ and the direction is implicitly defined according to the respective method.

\noindent\textit{k\_nearest\_neighbors}: 
The nearest neighbor function and its underlying metric have significant impact on the time efficiency and the theoretical properties of the algorithm. 
For curvature constrained instruments the Euclidean metric does not represent a good approximation on the actual distance. 
On the other hand, the computation of a more complex metric like the reachable set of a particular state \cite{shkolnik2009:reachabilityRRT} can be very time consuming. 
As the main interest in this application lies in the fast computation of a feasible path, 
we return the k-nearest neighbors in terms of the efficiently computable metric 
\[d(q_1(x,h_1),q_2(y,h_2)):\SEmath\times\SEmath\rightarrow\mathbb{R}^3 \coloneqq \Vert x-y\Vert_{\mathbb{R}^3} + \rho(h_1,h_2)\]

\noindent\textit{steer:} 
\MyAlgB~propagates the search along states on circular arcs. The local planner of \MyAlgA~uses a spline consisting of two Bézier-Spirals to expand the search tree. 
We refer to the original papers \cite{patil2014:rapidReplanning} and \cite{yang2014:SBRRT} for a detailed description.

\noindent\textit{attempt\_connection}: 
The original RRT-Connect does not address nonholonomic planning and considers the trees connected if both trees meet at the random sample. 
This approach would result in a discontinuous orientation at the connecting state as we sample only in $\mathbb{R}^3$ and do not enforce a specific orientation. 
Instead, a two point boundary value problem has to be solved in our approach to match both position and orientation: 

First, we search for a state of the other tree in the vicinity of $q_{next}$. Specifically, we check if a state lies within a cone that apex and direction is described by the location and orientation of $q_{next}$. If such a state is found, we try to connect these two:

The \MyAlgB~algorithm connects two corresponding states by solving the 3D Dubins problem with the geometric approach presented in \cite{hota2010:Optimal3D}. A similar method is used in \cite{pharpatara2017:trajectoryRRTstar}. Both papers address the computational complexity of their approach. However, our c++ implementation requires on average only 45 microseconds to solve the underlying nonlinear system of equations which makes it suitable for fast computation.

The \MyAlgA~algorithm iteratively uses the local steering function to steer from $q_{next}$ to its counterpart and vice versa. This procedure is repeated until either the interpolation criterion of the Bézier-Spline is satisfied during an iteration or the states missed each other and thus no connection was possible.

\section{Scenarios for the Temporal Bone}
\label{sec:data}
\begin{figure}
	\flushleft\introLegend
	\resizebox{\linewidth}{!}{%
		\includegraphics[height=1cm]{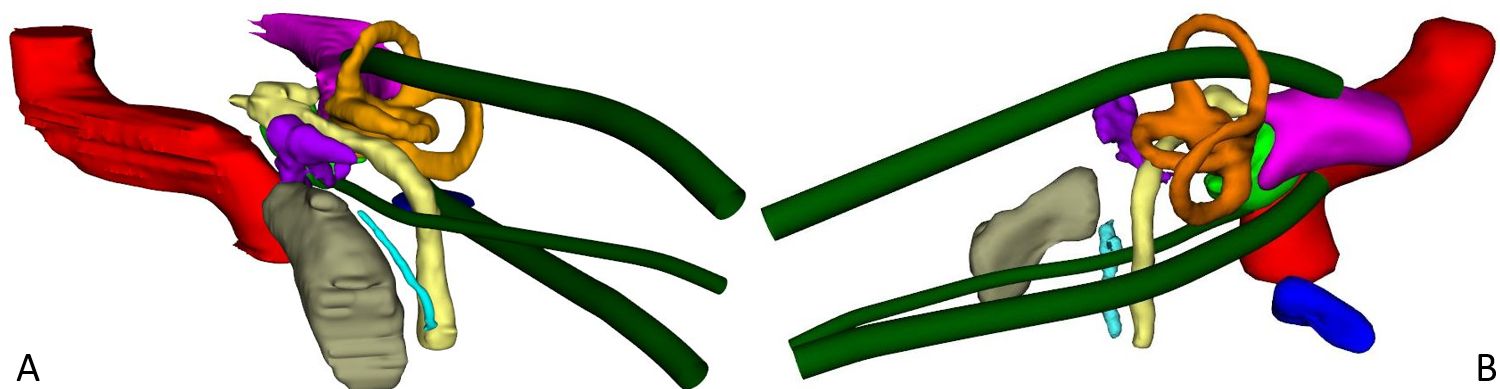}
	}
	\caption{
		Examples of the access paths (Cochlea-/SSC-/RL-Access) for real (\textbf{A}) and synthetic (\textbf{B}) anatomies.
	}
	\label{fig:samples}
\end{figure}

We address three typical medical interventions for the experiments to show the general suitability for temporal bone surgery (Fig. \ref{fig:samples}): 
one access to the cochlea via the facial recess (Cochlea-Access) and two accesses to the internal auditory canal: 
through the superior semicircular canal (SSC-Access) and via the retro-labyrinthine region (RL-Access). 
Parameters for each Problem Formulation (see Section \ref{sec:objective}) are listed in Table \ref{tab:params}. 
The curvature constraint reflects our current robot prototype. 
We tested with higher values of $r_d$ and $d_{max}$ for the RL- and SSC-Access as there is usually more space between obstacles. 
The time and orientation constraints were chosen according to real applications \cite{schulman2014:convexOptimization,torres2017:improvementCochleaAngle}.
\begin{table}[h]
	\centering
	\caption{Parameters for the Problem Formulations (see section \ref{sec:objective}) of the three access paths.}
	\label{tab:params}
	\begin{tabular}{l|ccccccc}
		& $\kappa_{max} $ & $\epsilon_G$ & $\phi_G$ (deg) & $d_{max}$ (mm) & $r_d$ (mm) & $T_{max}$ (s) \\
		\hline
		Cochlea-Access & 0.05 & 1.0 & 5 & 0.3 & 0.5 & 0.5 & \\
		SSC-Access     & 0.05 & 1.0 & 5 & 0.5 & 1.0 & 0.5 & \\
		RL-Access      & 0.05 & 1.0 & 5 & 1.0 & 1.0 & 0.5 & \\
	\end{tabular}
\end{table}

The risk structures, i.e. organs in the vicinity of access paths that must not be harmed, were extracted from real CT data of patients. 
To this purpose, our clinical partners manually segmented the internal carotid artery and jugular vein bulb, facial nerve and chorda tympani, cochlea, ossicles and labyrinth as well as the internal and external auditory canal in 40 high quality, but typical routine CT scans of the human temporal bone (Siemens Somatom, average resolution $0.18\times0.18\times0.4 mm^3$).

The manual assembly of such real scenarios is a necessary but extremely laborious and time consuming task. However, a statistical analysis of the motion planner's performance requires a much larger number of samples than this manual procedure can provide. 
Consequently, we divided the experiments into two setups:

\noindent\textbf{Real Anatomies: }
For the first 22 data sets we also segmented the brain and the skull's surface. In the resulting 3D environment, entry and target positions of potential interventions were manually placed in each individual data set with the help of a custom planning tool to provide samples on real patients (Fig. \ref{fig:samples} A).

\noindent\textbf{Synthetic Anatomies: }
First, we created statistical shape models \cite{cootes1995:asm} of the manually segmented risk structures of the otobasis in all 40 data sets. Then, we generated 100 synthetic anatomies based on the real ones. For each new synthetic anatomy, one of the real anatomies was chosen randomly to serve as an atlas, including its risk structures and its goal regions of the three potential interventions. A variation of the statistical shape models was then registered to the atlas to replace each original structure with an altered variant (Fig. \ref{fig:samples} B).

\section{Experiments}
\label{sec:experiments}
In the following, we describe in detail the setup of real and synthetic anatomies as well as the parameters of our motion planners.

\noindent\textbf{Real Anatomies:} 
\noindent In each data set and for each of the three applications (RL-, SSC-, Cochlea-Access) we placed one state within the temporal bone and one state on the skull's surface to define the regions $C_I$ and $C_G$ of the Problem Formulations. 
Start states were positioned at the bottom of the internal auditory canal, at its top and next to the round window for the RL-, SSC- and Cochlea-Access, respectively.
This resembles a potential position of an acoustic neuroma (RL-, SSC-Access) or the entry point of the electrode in a cochlear implant (Cochlea-Access). 
The directions at these start states were defined as a compromise between the respective organ's normal at this position and a direction towards the skull's surface. 
Last, three states were placed on the skull with orientations approximately orthogonal to its surface which serve as goal states for the individual access paths.

%\textit{Note:} In all cases, it is easier for the methods based on single RRTs to find a path from the cluttered region inside towards the surface instead of the other way around. This way the start region becomes the goal point or target of the actual intervention which in some explanations might sound misleading.

\noindent\textbf{Synthetic Anatomies:} 
\noindent For each new synthetic anatomy, random variations of the individual statistical shape models' modes were computed by sampling the corresponding eigenvalues between $\pm1.0$ times of their standard deviation. 
The resulting model was then registered with the reference atlas. 
For the respective goal states we used the ones in the atlas. 
The start states required a new strategy for positioning, as their original pose in the atlas might be invalid. 
Thus, new start states were placed above/below the center of mass of the internal auditory canal (SSC-/RL-Access) and below the center of mass of the cochlea (Cochlea-Access). 
For orientation, individual reference points $P_{ref}\in\mathbb{R}^3$ were computed: 
For the RL-Access slightly inferior to the lower side of the bounding box of the facial nerve; for the SSC-Access above the center of mass of the semicircular canals and 
for the Cochlea-Access in the center of mass of the chorda tympani. 
The start states were then oriented so that the z-axis of the local coordinate frame points to the respective reference point.

\noindent\textbf{Motion Planning:} 
In both setups we let each of the four planners of Section \ref{sec:methods} calculate as many paths as possible within 20 seconds for all three applications. 
We used the number of found paths to quantify the performance of each planner. 
In order to compare the quality of paths computed by each planner, we measured for each trajectory both the deviation at the goal state and the minimal distance to risk structures. 

For goal biasing we chose a value of $25\%$. 
The \textit{attempt\_connection} method of \MyAlgs~was most successful with parameters $h=5.0$\,mm and $\alpha=\ang{20}$ for height and angle of the cone. 
A kd-Tree was used for collision checking between states and obstacles.
All experiments were performed on a system with an Intel Core i5-6500 CPU @ 3.20 GHz and 32,0 GB RAM. \\

\section{Results}
\label{sec:results}
We start with analyzing the motion planners' results on real anatomies. 
Then, we discuss their generalization on synthetic data.

\begin{figure}[t]
	\resizebox{\linewidth}{!}{%
		\includegraphics[height=1cm]{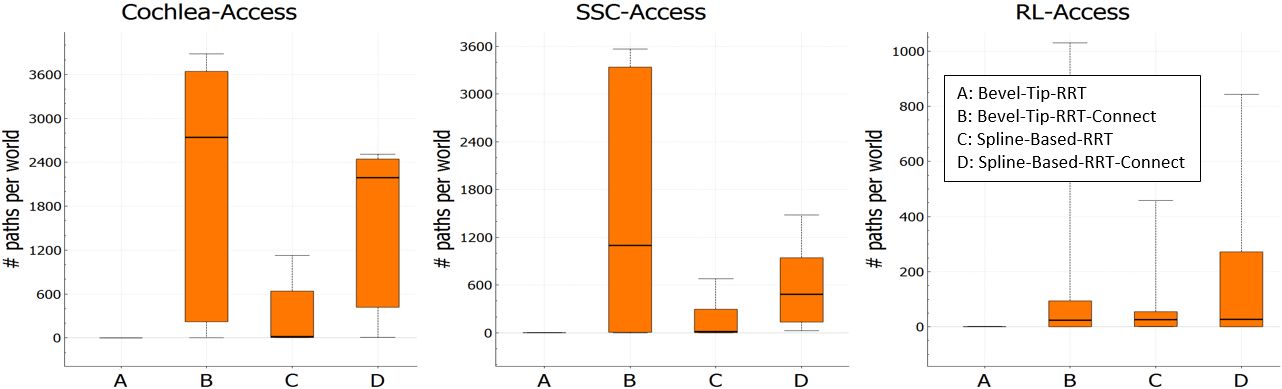}
	}
	\caption{
		Box-Plots for each access canal about the number of paths found by the individual planners in 22 real anatomies (higher=better).
	}
	\label{fig:boxplots_successrate_real}
\end{figure}

\noindent\textbf{Real Anatomies:} 
First, we look at the number of paths found in a specific time to ensure a planner is fast enough for intra-operative replanning \cite{patil2014:rapidReplanning}. 
Figure \ref{fig:boxplots_successrate_real} and Table \ref{tab:realSuccessRates} show the statistical distributions: 
For both the Cochlea- and the SSC-Access our RRT-Connect algorithms clearly outperformed standard RRT planners. 
In the case of the RL-Access the Spline-Based-RRT showed similar performance but none of the three algorithms really stands out. 
The number of paths found per second and the low number of failures indicate that \MyAlgs s work very well for the first two access canals and we can expect that successful intra-operational planning can be performed in minimal time.
In contrast, the search through the retro-labyrinthine region was unsuccessful for almost half of the anatomies. 
This is, however, not unexpected because the risk structures vary highly between patients: 
in case of a narrow passage between facial nerve and chorda tympani, a small semicircular canal or a high reaching bulb of the jugular vein, the creation of a feasible access path was impossible.
Indeed, a careful inspection showed that in the 6 cases algorithm C failed, a high reaching jugular vein bulb made a trajectory of the requested size completely impossible. 
The discrepancy between the first two problem formulations and the latter is also due the nature of relevant obstacles in the respective area. 
In the first two cases a bottleneck had to be passed (two nerves / the SSC), whereas for the RL-Access the facial nerve and the jugular vein had to be circumnavigated.

\begin{table}
	\caption{Performance of each planner for the real anatomies. Measured in median number of paths (\#), median number of paths per second (\#/s) and percentage of failed scenarios (F).}
	\label{tab:realSuccessRates}
	\centering
	\begin{tabular}{l | rrr | rrr | rrr}
					& \multicolumn{3}{c|}{Cochlea-Access} & \multicolumn{3}{c|}{SSC-Access} & \multicolumn{3}{c}{RL-Access}\\
					& \# & \#/s & F(\%) & \# & \#/s & F(\%) & \# & \#/s & F(\%) \\
		\hline
        Bevel-Tip (A)
		 & 0 & 0 & 80
		 & 1 & 0.05 & 50
		 & 0 & 0 & 75 \\
        Bevel-Tip-Connect (B)
		 & \textbf{2635} & \textbf{131.75} & 5
		 & \textbf{760} & \textbf{38} & \textbf{0}
		 & 4 & 0.2 & 45 \\
        Spline-Based (C)
		 & 17 & 0.85 & 5
		 & 14 & 0.7 & 5
		 & 9 & 0.45 & \textbf{25} \\
        Spline-Based-Connect (D)
		 & 2031 & 101.55 & \textbf{0}
		 & 442 & 22.1 & \textbf{0}
		 & \textbf{17} & \textbf{0.85} & 40 \\
	\end{tabular}
\end{table}

\begin{figure}[b]
	\resizebox{\linewidth}{!}{%
		\includegraphics[height=1cm]{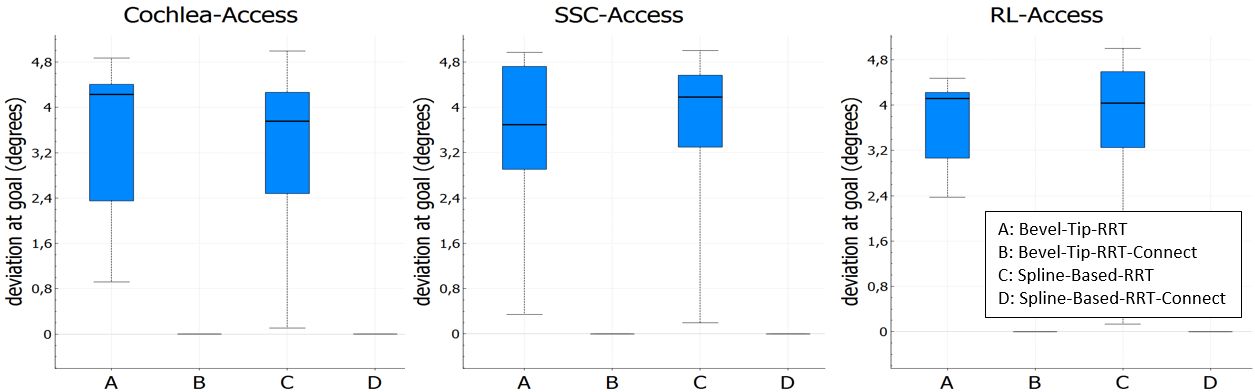}
	}
    \caption{Box-Plots about the deviation at the goal for the real anatomies (lower=better). }

	\label{fig:boxplots_angles_real}
\end{figure}

Now we look at the matching of the goal's pose. 
Naturally, RRT-Connect algorithms matched the orientation of goal states perfectly, whereas the RRTs were limited to an approximation (Fig. \ref{fig:boxplots_angles_real}). 
We also note, that in all three cases both \AlgBRRTfull~and \AlgSBRRTfull~tended to accomplish the maximal allowed deviation rather than a perfect match.

\begin{figure}
	\centering
	\textbf{A} Bevel-Tip-RRT, \textbf{B} Bevel-Tip-RRT-Connect - \textbf{C} Spline-Based-RRT - \textbf{D} Spline-Based-RRT-Connect\\
	\resizebox{\linewidth}{!}{%
		\includegraphics[height=1cm]{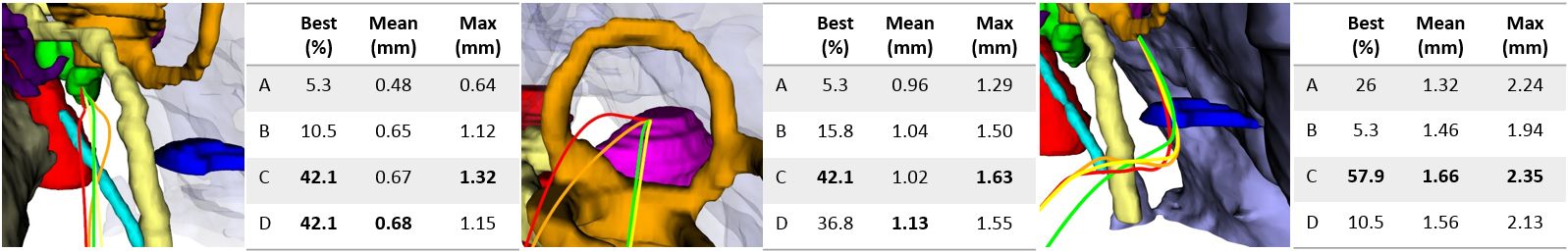}
	}
	\caption{Close-Up of the narrowest region of each access canal. The corresponding table shows the mean and max distance of each planner over all real anatomies together with the percentage of how often each planner found the best path according to the maximum distance.
	}
	\label{fig:statisticsDistances}
\end{figure}

Next, we focus on the minimal distance an access path had to risk structures as this is usually the most relevant metric to clinicians. 
To this purpose, we interpolated between the states of the search tree at a resolution of 0.1\,mm. 
For each of those interpolated states, we then sampled points on a circle with radius $r_d$ and orthogonal to the state's direction and computed the minimal distance to the next obstacle.
Figure \ref{fig:statisticsDistances} shows in small images the narrowest region that had to be passed together with three statistics for each planner across all 22 anatomies: 
the percentage how often it computed the best path for a specific anatomy (Best), the mean minimal distance its best path had to risk structures (Mean) and the overall best path it computed across all anatomies (Max). 
Clear superiority of a specific algorithm was not observable although the \AlgSBRRTfull~tended to find paths with the largest distance more often. 
Hence, our new \MyAlgs~did not suffer from lower quality. 
From the observed distances we also got an impression of the average size of the passed bottleneck. 
This can help in the design for the robot prototype. 
According to Table \ref{tab:realSuccessRates}, for example, \MyAlgs~always found trajectories for an SSC-Acces with the specifications in Table \ref{tab:params}, having on average still a minimal distance above 1.0\,mm to the nearest obstacle.

\begin{figure}[t]
	\resizebox{\linewidth}{!}{%
		\includegraphics[height=1cm]{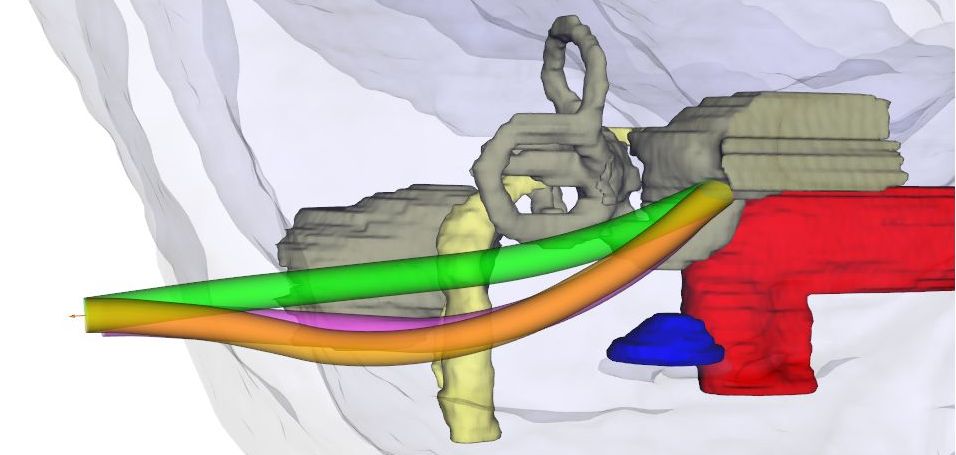}
		\includegraphics[height=1cm]{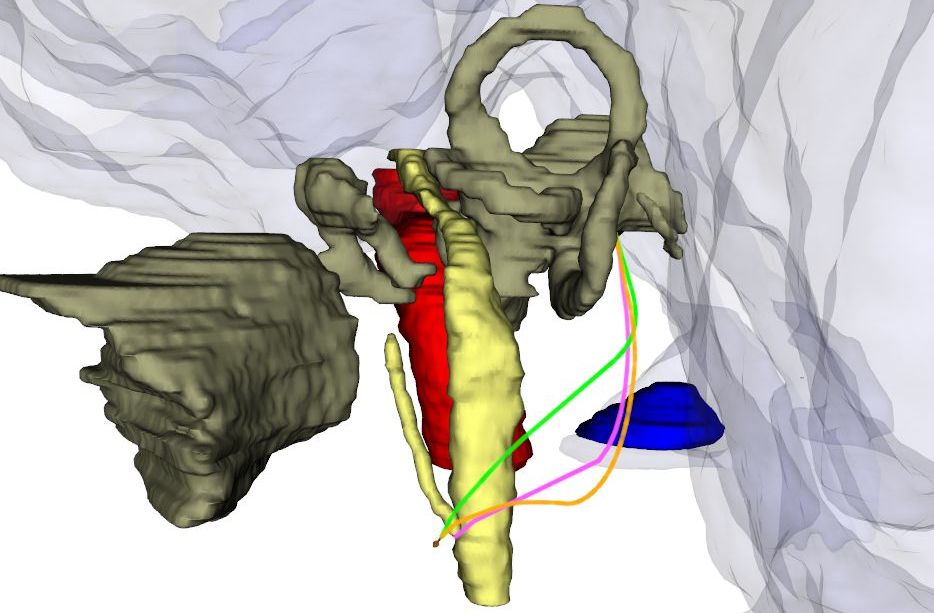}
	}
	\caption{
		RL-Access planned by a standard RRT (pink tube) with safety distance 1.0\,mm and by a \MyAlgs~(green, orange) with safety distances 1.0\,mm and 1.5\,mm, respectively.
		}
	\label{fig:sampleHigherSafetyDist}
\end{figure}

Last we address the issue that in many scenarios the \AlgSBRRTfull~found paths with the highest minimal distance. 
A closer inspection showed that the \MyAlgs~just quickly found a solution as soon as the relevant obstacle had been passed. 
When we enlarged the allowed safety distance, the \MyAlgs~computed paths with similar minimal distances.
Figure \ref{fig:sampleHigherSafetyDist} shows an example of this behavior for the RL-Access with safety distance 1.0\,mm and 1.5\,mm.

\noindent\textbf{Synthetic Anatomies:} 
To study the generalization of these specific cases, we then looked at synthetic scenarios. Instead of real anatomies we now worked with variances based on atlases of real data combined with the shape space of the statistical shape models.
Our evaluations then included a much broader variety of anatomies. 
By randomly sampling the shape space we also made sure, that the individual real anatomies did not provide edge cases for the algorithms, a standard approach in motion planning \cite{karaman2012:holonomicAlgorithms,gammell2014:BIT}.

\begin{figure}[t]
	\resizebox{\linewidth}{!}{%
		\includegraphics[height=1cm]{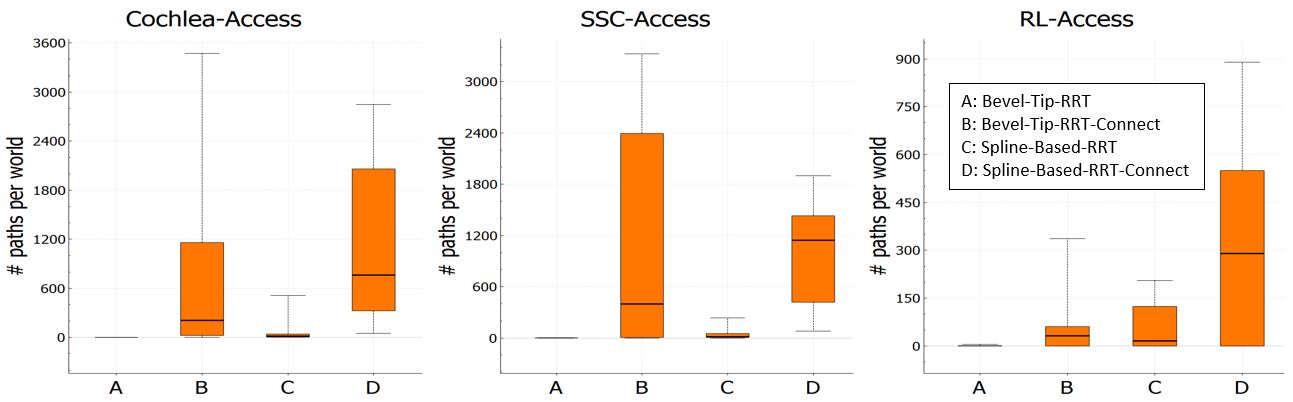}
	}
	\caption{
		Box-Plots about the success rates of the planners in 100 synthetic anatomies (higher=better). 
	}
	\label{fig:boxplots_successrate_random}
\end{figure}

\begin{figure}[t]
	\resizebox{\linewidth}{!}{%
		\includegraphics[height=1cm]{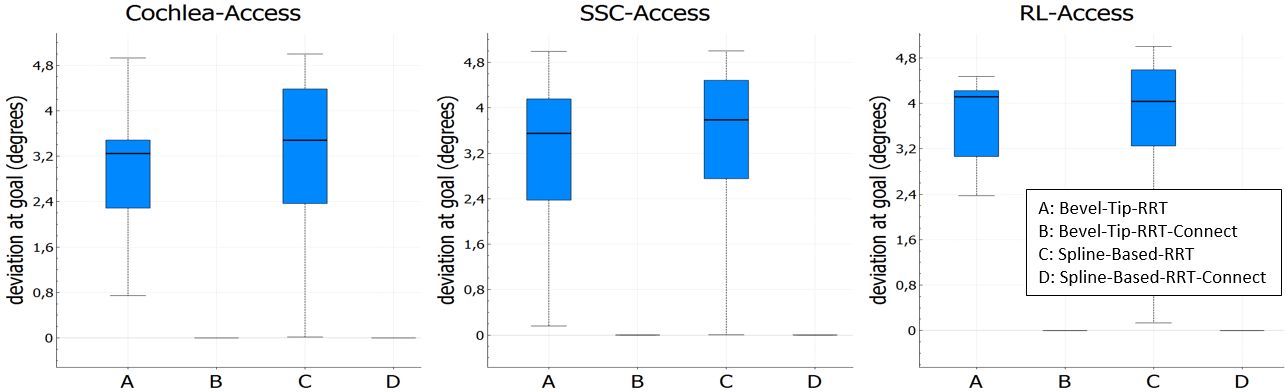}
	}
	\caption{
		Box-Plots about the deviation at the goal for 100 random synthetic anatomies (lower=better). 
	}
	\label{fig:boxplots_angles_random}
\end{figure}

The results in Figure \ref{fig:boxplots_successrate_random} and Table \ref{tab:syntheticSuccessRates} show how the planners performed for each access canal. 
From the box plots we can conclude, that \MyAlgs~again tended to find many more paths. 
Their performances according to Table \ref{tab:syntheticSuccessRates} supported the results of the real cases: 
For the given parameters of Table \ref{tab:params} access paths for the Cochlea- and SSC-Access were always possible whereas for the RL-Access a high reaching jugular vein often prevented a feasible trajectory to be found.
The number of paths found per second again indicated that bidirectional RRTs are suitable for intra-operational planning.
An analysis of the orientation at the goal showed equivalent results to the real cases: 
RRTs hardly realize a good match of the desired orientation (Fig. \ref{fig:boxplots_angles_random}). Although this was expected, it clearly supports our claim, that bidirectional planners are required, if precise replanning is necessary.

%Last we compare the observed minimal distances according to Figure \ref{fig:synthetic_closeUps}. 

\section{Conclusion}
\label{sec:conclusion}
In this paper, we address a minimally-invasive procedure with demands on fast computation and high precision of both initial and goal pose. 
We present two suitable RRT-Connect motion planners, one based on Bézier-Splines, the other on circular arcs and 3D Dubins Paths, which quickly compute feasible curvature constrained access paths for the proposed interventions. 
The efficiency of these planners is shown in real CT data of patients as well as on randomized anatomies created from variations of statistical shape models. 
These tailored RRT-Connect algorithms outperform state of the art one-directional planners and provide a reliable and fast method for planning access paths in temporal bone surgery.

%The efficiency of these planners was shown in real CT data of patients and novel randomized worlds created from variations of statistical shape models. 
%The methods were also compared to existing RRT-versions and a significant improvement was shown. 
%Therefore, the approach with tailored RRT-Connect algorithms provides a reliable and fast method for planning access paths in temporal bone surgery.

In the future we want to improve the approach for both methods with an optimization of planned paths regarding larger distances to risk structures or more advanced metrics. 
We also expect that an improvement of the connection method of our RRT-Connects will result in better performances for difficult cases like passing the retro-labyrinthine region.
Moreover, we would like to investigate the applicability of these general purpose planners for other medical interventions such as needle insertion in soft tissue \cite{schulman2014:convexOptimization} or flexible endoscopes \cite{fichera2017:ThroughTheTube}. 
We believe such precise nonlinear planning procedures are expected to be instrumental in improving interventions and advancing patient safety at operating rooms of the future. 

\begin{table}
	\centering
    \caption{
    	Measured in median number of paths (\#), median number of paths per second (\#/s) and percentage of failed scenarios (F).
    }
	\label{tab:syntheticSuccessRates}
	\begin{tabular}{l| rrr | rrr | rrr}
		& \multicolumn{3}{c|}{Cochlea-Access} & \multicolumn{3}{c|}{SSC-Access} & \multicolumn{3}{c}{RL-Access}\\
		& \# & \#/s & F (\%) & \# & \#/s & F (\%) & \# & \#/s & F (\%) \\
		\hline
		Bevel-Tip (A)
		 & 0 & 0 & 80
		 & 0 & 0 & 57 
		 & 0 & 0 & 66 \\
        Bevel-Tip-Connect (B)
	     & 208 & 10.4 & \textbf{0}
	     & 398 & 19.9 & 12
	     & 27 & 1.35 & 37 \\
        Spline-Based (C)
	     & 15 & 0.75 & 7
	     & 14 & 0.7 & 7
	     & 15 & 0.75 & \textbf{26} \\
	     Spline-Based-Connect (D)
	     & \textbf{762} & \textbf{38.1} & \textbf{0}
	     & \textbf{1144} & \textbf{57.2} & \textbf{0} 	
	     & \textbf{273} & \textbf{13.65} & 30 \\
	\end{tabular}
\end{table}

\begin{figure}[t]
	\resizebox{\linewidth}{!}{%
		\includegraphics[height=1cm]{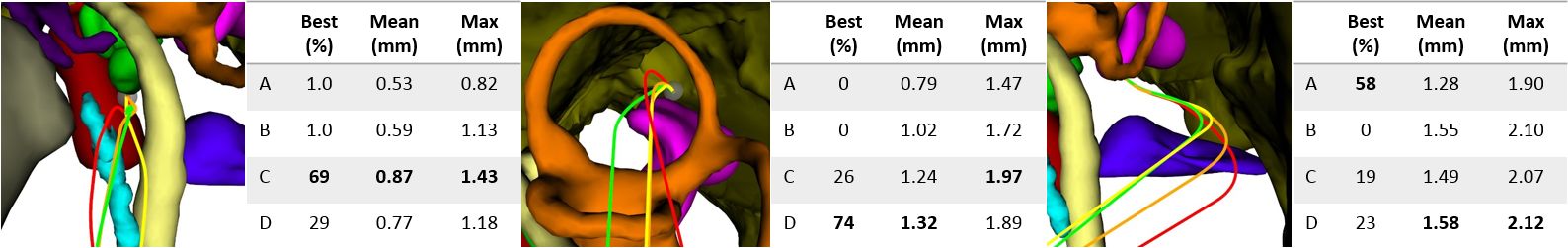}
	}
	\caption{
		Close-Up of the narrowest region of each access canal. The corresponding table shows the mean and max distance of each planner over all synthetic anatomies together with the percentage of how often each planner found the best path according to the maximum distance.
	}
	\label{fig:synthetic_closeUps}
\end{figure}

%\begin{acknowledgements}
%\section*{Acknowledgment}
\section*{Compliance with Ethical Standards}\label{CES}

\textbf{Disclosure of potential conflicts of Interest: }
The research project MUKNO II is funded by the DFG. The authors declare that they have no conflict of interest.

\noindent\textbf{Research involving Human Participants and/or Animals: }This article does not contain any studies with human participants or animals performed by any of the authors.

%\textbf{\\* Informed consent \\*} 
\noindent\textbf{Informed consent: }This article is partially based on anonymized patient data.
%\end{acknowledgements}

\bibliographystyle{spbasic}      % basic style, author-year citations
\bibliography{literature}   % name your BibTeX data base

%\begin{thebibliography}{}
% etc
%\end{thebibliography}

\end{document}